\documentclass[conference]{IEEEtran}
\IEEEoverridecommandlockouts
\usepackage{cite}
\usepackage{amsmath,amssymb,amsfonts}
\usepackage{algorithmic}
\usepackage{graphicx}
\usepackage{textcomp}
\usepackage{xcolor}
\usepackage{multirow}
\usepackage[breaklinks=true,colorlinks,bookmarks=false]{hyperref}
\def\BibTeX{{\rm B\kern-.05em{\sc i\kern-.025em b}\kern-.08em
    T\kern-.1667em\lower.7ex\hbox{E}\kern-.125em}}

\usepackage{eso-pic}

\begin{document}

\title{Influence of Segmentation on Deep Iris Recognition Performance\\
\thanks{
\textcopyright 2019 IEEE. Personal use of this material is permitted. Permission from IEEE must be obtained for all other uses, in any current or future media, including reprinting/republishing this material for advertising or promotional purposes, creating new collective works, for resale or redistribution to servers or lists, or reuse of any copyrighted component of this work in other works.
\noindent
DOI: \href{https://doi.org/10.1109/IWBF.2019.8739225}{https://doi.org/10.1109/IWBF.2019.8739225}
}
}

\author{\IEEEauthorblockN{
Ju\v{s} Lozej\IEEEauthorrefmark{1}, Dejan \v{S}tepec\IEEEauthorrefmark{2}, Vitomir \v{S}truc\IEEEauthorrefmark{3}, Peter Peer\IEEEauthorrefmark{1}}
\IEEEauthorblockA{\IEEEauthorrefmark{1}Faculty of Computer and Information Science, University of Ljubljana, Slovenia\\
E-mail: jlozej@gmail.com, peter.peer@fri.uni-lj.si}
\IEEEauthorblockA{\IEEEauthorrefmark{2}XLAB d.o.o., Ljubljana, Slovenia\\
E-mail: dejan.stepec@xlab.si}
\IEEEauthorblockA{\IEEEauthorrefmark{3}Faculty of Electrical Engineering, University of Ljubljana, Slovenia\\
E-mail: vitomir.struc@fe.uni-lj.si}
}

\maketitle

\begin{abstract}
Despite the rise of deep learning in numerous areas of computer vision and image processing, iris recognition has not benefited considerably from these trends so far. Most of the existing research on deep iris recognition is focused on new models for generating discriminative and robust iris representations and relies on methodologies akin to traditional iris recognition pipelines. Hence, the proposed models do not approach iris recognition in an end-to-end manner, but rather use standard heuristic iris segmentation (and unwrapping) techniques to produce normalized inputs for the deep learning models. However, because deep learning is able to model very complex data distributions and nonlinear data changes, an obvious question arises. How important is the use of traditional segmentation methods in a deep learning setting? To answer this question, we present in this paper an empirical analysis of the impact of iris segmentation on the performance of deep learning models using a simple two stage pipeline consisting of a segmentation and a recognition step. We evaluate how the accuracy of segmentation influences recognition performance but also examine if segmentation is needed at all. We use the CASIA Thousand and SBVPI datasets for the experiments and report several interesting findings. 
\end{abstract}


\section{Introduction}
Iris recognition has for a long time been one of the most accurate and robust means of automated person identification. With a research history of more than twenty years and important applications in access control, banking, consumer electronics and forensics, it also represents one of the most mature branches of biometric recognition technology.

Iris recognition systems are typically composed of two main components: \textit{i)} an iris segmentation procedure that extracts the region-of-interest (ROI), i.e., the iris area, from the input image and maps the circular iris texture from a polar coordinate system to a Cartesian one using the rubbersheet model, and \textit{ii)} an feature extraction (and recognition) procedure that encodes the normalized texture into a descriptive and discriminative feature vector that is then used for similarity measurements and identity inference. Multiple techniques have been presented in the literature for both the segmentation as well as the texture-representation step and we refer the reader to~\cite{nigam2015ocular} for more information on existing approaches.

With the recent success of deep learning in different areas of computer vision, research in iris recognition is also starting to look at deep learning methodologies. A number of solution has been  presented recently in the literature for iris segmentation, e.g.,~\cite{rot2018deep,lozej2018end,jalilian2017domain,arsalan2017deep,jalilian2017iris,yang2018robust,koh2010robust}, and recognition~\cite{deepIris,lwIris,offTheShelf,iccv17}. However, these typically do not approach iris recognition in an end-to-end manner, but either segment the iris using supervised deep learning models and then represent the iris texture using established iris encoding techniques, e.g.,~\cite{hofbauer2018exploiting}, or first segment the iris from the input image using standard iris-segmentation approaches and then process the unwrapped iris texture using deep learning models, e.g.,~\cite{deepIris}. While the interaction of deep segmentation models and standard iris representation techniques has been studied recently~\cite{hofbauer2018exploiting,kinnison2019learning}, the influence of segmentation techniques on the performance of deep learning models has, to the best of our knowledge, not been investigated so far for the task of iris recognition.

In this paper, we try to address this gap and analyze the impact of iris segmentation in a deep-learning-based iris recognition system. We use an off-the-shelf convolutional neural network (CNN) recognition model to encode the iris texture and perform  various experiments aimed at assessing the impact of the segmentation procedure on the iris recognition performance. Specifically, we perform experiments: \textit{i)} with automatically generated segmentation masks using CNN-based segmentation models, \textit{ii)} manually annotated ground-truth segmentations, but also \textit{iii)} without any segmentation masks at all. Because deep models are able to model complex (nonlinear) data changes, we skip the iris-unwrapping step altogether and show that highly competitive performance can be achieved even without normalizing the iris texture. For comparison purposes we also include results for standard (heuristic) segmentation techniques, followed by iris unwrapping. We conduct experiments on the CASIA-Thousand (containing near infra-red (NIR) images) and SBVPI (containing visible spectrum (VIS) images) datasets and as a side product of our analysis also show that a single CNN-based recognition model can be used to encode heterogeneous iris textures captured in different parts of the EM spectrum.

In summary, we make the following main contributions in this paper:
\begin{itemize}
    \item We present an analysis of the impact of iris segmentation on the recognition performance of deep learning models.
    \item We show that a single deep learning model can be used to efficiently encode iris texture in both, the near infra-red as well as the visible spectrum.
    \item We make all models, weights and source code publicly available via {\small https://github.com/jus390/segInfluence}
\end{itemize}





The rest of the paper is structured as follows: In Section~\ref{Sec: Related work} we briefly review the related work of relevance to this paper. In Section~\ref{Sec: Methods} we describe the model we used for our evaluation and procedure used to learn the model parameters. In Section~\ref{Sec: Results} we present the results of our experiments and discuss our main findings. Finally we conclude the paper in Section~\ref{Sec: Conclusion}.

\section{Related work}\label{Sec: Related work}

As indicated above, deep iris recognition is still a developing research area and not many solutions have been presented in the literature so far. In this section, we cover only existing work related to deep iris recognition and refer the reader to~\cite{rot2018deep,lozej2018end,hofbauer2018exploiting,kinnison2019learning} for more information on  deep models for other iris-related problems, such as segmentation. 

The first deep iris recognition approach was presented in 2016 \cite{deepIris}. Here, the authors, Gangwar and Joshi, presented two architectures, collectively called DeepIrisNet. The first uses a simple VGG-like \cite{vgg} structure, while the second replaces the final two blocks with inception modules \cite{incept}. The authors identify individuals based on the features obtained from the final fully connected layer, which are compared using the Euclidean distance. 
In \cite{lwIris} a light weight architecture with a much smaller amount of training parameters is presented. Here, the authors capitalize on use of texture-like information, favoring features from the convolutional layers instead of the fully connected layer. The authors binarize features using an ordinal measure and compare feature vectors using the Hamming distance. 
In \cite{iccv17}, Zhao and Kumar present a pipeline called UniNet, which consist of two sub-networks: MaskNet and FeatNet. Both networks use preprocesed normalized iris images. FeatNet is used to extract features encompassing different levels of detail, while MaskNet removes details not belonging to the iris. The authors also introduce a specialized loss function called extended triplet loss, which directs training towards more texture-like information. 
Finally, Nguyen et al. \cite{offTheShelf} suggest using existing readily available (also referred to as off-the-shelf) models for recognition. The authors test a number of architectures, while also studying which layers ensure the best results. The extracted features are compared using a support vector machine (SVM).

The majority of approaches described above uses traditional techniques to pre-segment and pre-normalize images before feeding them to the CNN-models. Thus, these models still follow standard iris processing pipelines, where an iris representation is computed from segmented and unwrapped iris textures. Here, we consider a different approach, where both segmentation and iris-representation steps are implemented using deep models, and study the impact of segmentation on iris recognition performance in such a setting. 
\section{Methods}\label{Sec: Methods}
In this section we describe the methodology used for our analysis. We first discuss the deep learning framework  used for our experiments and then elaborate on the training procedure used to learn the parameters of the experimental processing pipeline.

\subsection{Deep learning framework}
\begin{figure*}[htbp]
\centerline{\includegraphics[width=0.85\textwidth]{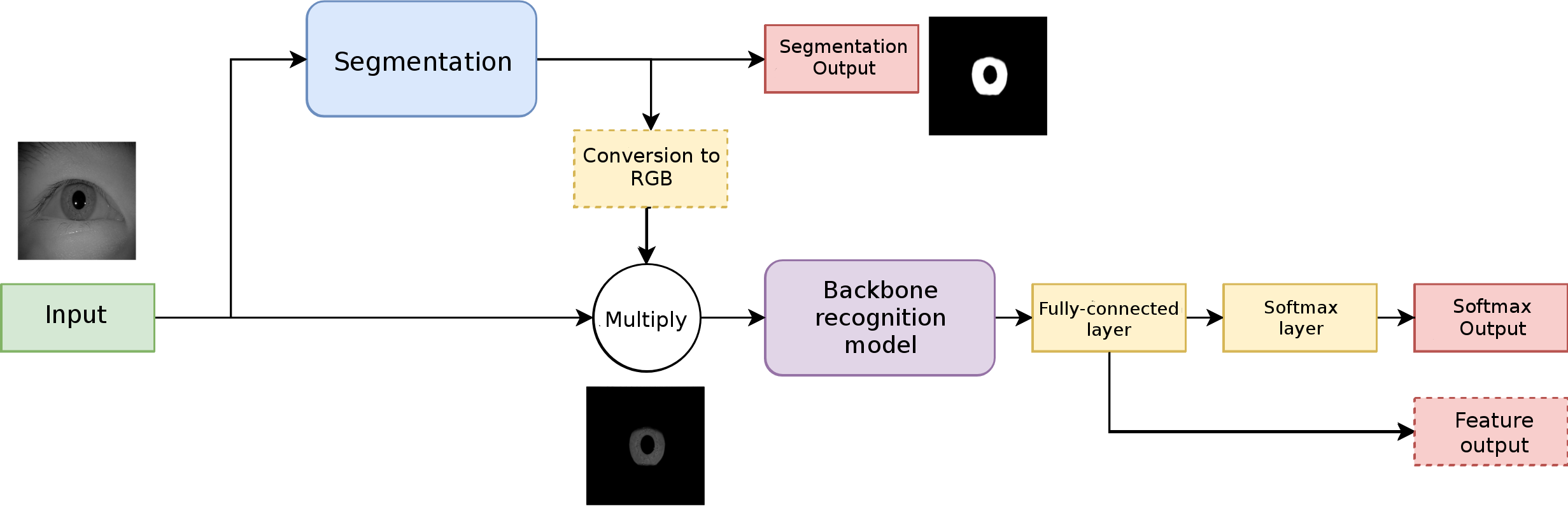}}
\caption{Illustration of the deep learning framework used in our analysis. The framework consists of two main steps: a segmentation and a recognition (or encoding) step. The segmentation step extracts a mask of the iris region from the input image and uses the segmentation result to mask out all image pixels not belonging to the iris. To achieve this the generated segmentation mask is multiplied with the input image. The masked iris region is fed to a deep (iris recognition) model that is used to extract features for the recognition experiments.}
\label{fig: arh}
\end{figure*}

For the experiments, we use a simple two-stage recognition pipeline consisting of a segmentation step and a recognition step, as illustrated in Fig. \ref{fig: arh}. The pipeline is straight-forward and very flexible, which allows us to easily replace components or introduce manually annotated segmentation masks into the experimental setup. 

As can be seen in Fig. \ref{fig: arh}, input images are first fed into a segmentation model, which extracts a mask corresponding to the iris area, i.e., the ROI. This mask is then multiplied with the initial input image, masking out (i.e. multiplying them with zero) image regions, which are not part of the iris. The segmentation step and corresponding masking procedure can be seen as a form of attention mechanism for the recognition model that forces the model to learn only iris-related features and to ignore other parts of the image, such as the periocular region. The masked image is finally fed into the recognition part (or better said, feature extraction part) of the pipeline, which computes features for the identification procedure. 

In our implementation we utilize a DeepLabV3+ \cite{deepLabv3plus} with a MobileNet \cite{moblienet} backbone (and an output stride of $16$) for segmentation and an Xception \cite{xception} model for recognition. We decided to use these particular models because they give good results on a number of vision-realted tasks and feature a relatively small amount of  parameters that need to be learned during training. Our final implementation has $29,494,217$ open parameters, of which $2,196,033$ come from the segmentation model, and $27,298,184$ come from the recognition part. Our final models are roughly 320 MB large.

\subsection{Training} 

In order to train the framework described above, ground truth segmentation masks as well as identity labels are need for all training data. Because only a small subset of the training images we use (see Section~\ref{subsec: training data} for details) has annotated segmentation masks, we trained our model in two steps. The first step is dedicated to learning the parameters of the segmentation part of the pipeline, and the second to learning the parameters of the recognition model. 

To learn the parameters of the segmentation model and generate segmentation results that are best suited for the subsequent recognition procedure, we trained both the segmentation part of the pipeline as well as the the recognition part in conjunction. We achieve this by using outputs of both models (segmentation and recognition) each with its own loss function. Binary cross entropy is used as the loss over the segmentation outputs and categorical cross entropy is utilized as the loss for the recognition model. The final loss function is a weighted sum of both, but with a greater weight on the segmentation loss. When setting weights for the losses, we pay attention to normalize each function so that they give results in a similar numeric range. Note that the described procedure is a standard end-to-end training procedure, which, to the best of our knowledge, has not been used for iris recognition so far. Here, we use it as a training step for the segmentation model only, as we have limited training data with the needed ground truth segmentation masks available. Because the amount of such data is not sufficient to learn competitive recognition models, we use a secondary learning step to learn the parameters of the recognition models. However, with a sufficient amount of training data, the entire model could be trained end-to-end.

Once training of the segmentation part is completed, we continue with the second step of training. In this step we train the recognition part on a larger amount of samples (see Section~\ref{subsec: training data} for details). To make sure that the recognition is trained only on the iris region, we freeze the layers of the segmentation model, thus preventing changes in the segmentation results.
Once the entire pipeline is trained, we use features from one of the  layers of the recognition models for comparing iris images - Fig.~\ref{fig: arh} illustrates the feature extraction step for the fully connected layer of the recognition model.

In both training steps augmentations are utilized in order to increase the amount of available training data and avoid over-fitting. We utilize variations in scale, rotation and location in order to diversify our data and regularize the training. In both steps we use the Adam optimizer \cite{adam} with a learning rate of $10^{-4}$ for optimization. We initialize the model using random weights. We conducted training until we see no improvements in the validation loss for $5$ training epochs. 
Segmentation training took around $30$ minutes to complete, while recognition was trained for roughly $10-12$ hours on a desktop PC with an Intel I7 2600k processor and an NVIDIA GTX 1060 GPU with 8 GB of RAM.

\section{Experiments and Results}\label{Sec: Results}

In this section we present the results of our experiments. We first discuss the dataset and protocol used for the experiments and then comment on the results of our assessment.

\subsection{Dataset and experimental protocol}\label{subsec: training data}

We select two datasets for our experiments, i.e., the CASIA-Iris-Thousand \cite{casia} and SBVPI \cite{sbvpi} datasets.

CASIA-Iris-Thousand dataset contain 20,000 images of 1000 subjects.
Images from the CASIA dataset were captured using a near-infrared camera, and show forward gazing with and without glasses. The SBVPI dataset contains 3,708 images of 110 subjects.  Images from this dataset were captured in the visible-light spectrum so they also contain color information. Other than that, the eyes in SBVPI also exhibit large variations in gaze direction, e.g. each subject has images looking up, left, right and forward. Example images from both datasets can be seen in the first row of Fig. \ref{fig: seg}. The CASIA and SBVPI images have different resolution, so we resize all images to a fixed size of $320 \times 320$ pixels. 

Using the two datasets introduced above, we construct three datasets for training our deep iris recognition pipeline: two containing only samples from the experimental datasets and one constituting a balanced set of images from both datasets. The first training dataset (referred as \textit{CASIA} hereafter) comprises data from the CASIA-Iris-Thousand dataset and contains $14,000$ training images, $4,000$ validation images and $2,000$ images that were left out for further testing. The training images correspond to all $1,000$ individuals. $220$ of these images are used to train the segmentation part of our pipeline, and are further split between the actual training ($154$ images) and the validation data ($66$ images). The second training dataset (\textit{SBVPI} hereafter) contains $1,422$ training and $656$ validation samples belonging to $60$ individuals from the SBVPI dataset. Out of the available SBVPI training data, $332$ image are use for training the segmentation part of our pipeline and $58$ of these are used for validation. The third training set (referred to as \textit{COMBINED} hereafter) used images from both experimental datasets. The combined dataset contains $2,906$ training samples and $1,080$ validation samples. The combined segmentation dataset is constructed from all images with annotated ground truth segmentation masks of both segmentation datasets. 

Our testing dataset consists of $552$ annotated iris images corresponding to $23$ individuals from both CASIA-Iris-Thousand\cite{casia} and SBVPI dataset. The test set is kept constant for all experiments and contains heterogeneous images captured in the near-infrared (NIR) spectrum as well as iris images captured in the visible-light (VIS) part of the EM spectrum. The variability in our test set is, hence, across image characteristics (VIS vs. NIR), presence of glasses, gender, ethnicity, gaze direction, etc.  The experimental dataset is constructed out of $220$ images from CASIA belonging to $11$ individuals, while the other $332$ were from SBVPI belonging to $12$ individuals. Segmentation was tested on the validation subset of each dataset in order to utilize as many samples as possible.
\begin{table}[tb]
\caption{Segmentation results achieved with the trained segmentation model on the validation data.}
\centering
\begin{tabular}{| l | l | r | r |}
\hline

\textbf{Validation set}  & \textbf{Training set}& \textbf{Precision} [in \%] & \textbf{Recall} [in \%]  \\ \hline
CASIA  & CASIA & $91.79  \pm 3.05$ & $95.82 \pm 2.32$  \\ \hline
CASIA  & SBVPI & $6.42  \pm 1.53$ & $99.89 \pm 0.29$  \\ \hline
SBVPI  & SBVPI & $88.26 \pm 6.44$ & $90.57 \pm 7.00$  \\ \hline
SBVPI  & CASIA & $70.07  \pm 19.22$ & $63.07 \pm 33.24$  \\ \hline
CASIA & Combined & $91.83 \pm 3.04$ & $95.54 \pm 2.41$  \\ \hline
SBVPI & Combined & $93.27 \pm 4.68$ & $94.27 \pm 4.12$  \\ \hline
CASIA + SBVPI  & Combined & $91.97 \pm 4.14$ & $95.44 \pm 3.08$  \\ \hline
\end{tabular}
\label{tab:seg}
\end{table}


To identify individuals we compare features  from the global average pooling layer at the top of the recognition backbone. The reason for selecting the global average pooling layer is the better performance observed during our preliminary experiments, compared to the features from the fully connected layer. For feature comparisons the cosine similarity is used. 

For the analysis we conduct identification experiments (one vs. many) in an all-vs-all experimental protocol. Thus, each image in the test set is compared against a gallery comprised of the the entire test set. Self-comparisons are excluded from the results. We report results in terms of rank-1 and rank-5 identification accuracy and also plot Cumulative Match Characteristic (CMC) curves to provide more detailed insight into the experiments.

\subsection{Segmentation experiments}

To be able to analyze the influence of the segmentation procedure on the recognition performance of our deep iris recognition pipeline, we require an estimate of the segmentation performance of our model. Since we do not conduct explicit experiments towards this end, we report the precision and recall values achieved with our model on the validation data during the training procedure. Note that the validation data is not used directly to learn the parameters of the segmentation model, but only to determine when to stop training.

\begin{figure}[tb]
\centerline{\includegraphics[width=0.45\textwidth]{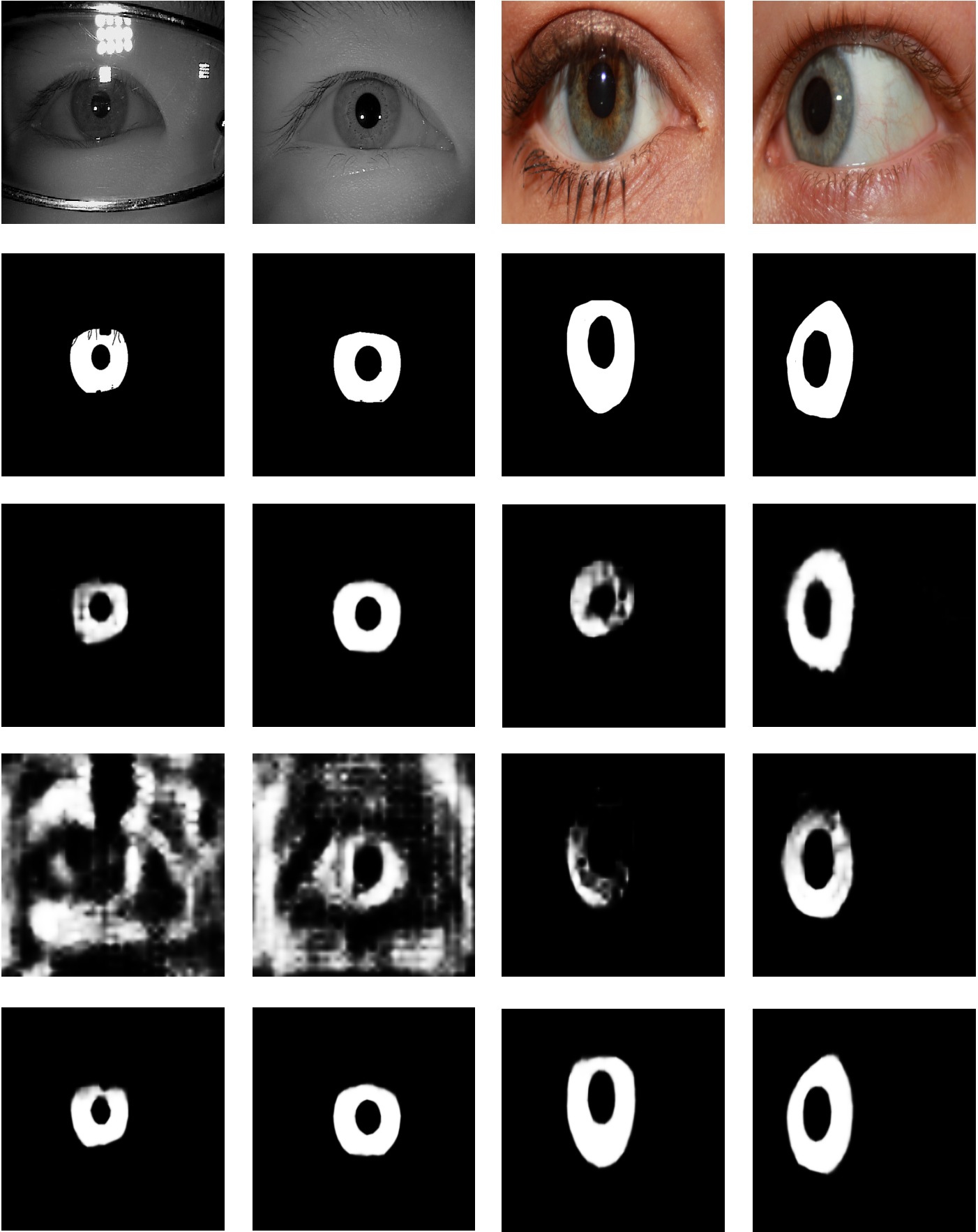}}
\caption{Examples of samples and their segmentations. In the first row we present the input images followed by their manual annotations. The third row contains the segmentation results of the models trained only on samples from the matching dataset. The next row contains the results of segmentation computed with the model trained on samples of the opposite model, e.g. a sample from CASIA was used in the model trained on SBVPI and vice versa. Finally the fifth row contains the segmentation results for the model that was trained on the combined training set containing images from CASIA and SBVPI. Note that despite the heterogeneous nature of the images, a single model is capable of ensuring good segmentations for both types of images.}
\label{fig: seg}\vspace{-2mm}
\end{figure}

The results in Table~\ref{tab:seg} and Fig.~\ref{fig: seg} show that the segmentation model perform extremely well when data with the same characteristics is used training and validation, e.g., NIR images are used for training and validation. In this setting average precision and recall values of close to (and over) 90\% are observed. Especially for the non-ideal iris images from the SBVPI dataset captured under different gaze directions these results are very competitive compared to traditional segmentation approaches. If the model is trained on NIR images and tested on VIS images (row 5 in Table~\ref{tab:seg}) or the other way around (row 3 in Table~\ref{tab:seg}), the performance degrades significantly. The best overall results on both types of images (NIR and VIS) are achieved if the combined dataset is used for training. Note that this result is not self-evident as the validation images of the two datasets are considerably different in terms of characteristics. To the best of our knowledge, this work is the first to show that a single deep segmentation model can be used to segment heterogeneous iris images. 

\subsection{Impact of segmentation}

\begin{figure*}[tb]
\begin{minipage}{0.325\textwidth}
\centerline{\includegraphics[width=0.995\textwidth,trim={5mm 3mm 15mm 5mm},clip]{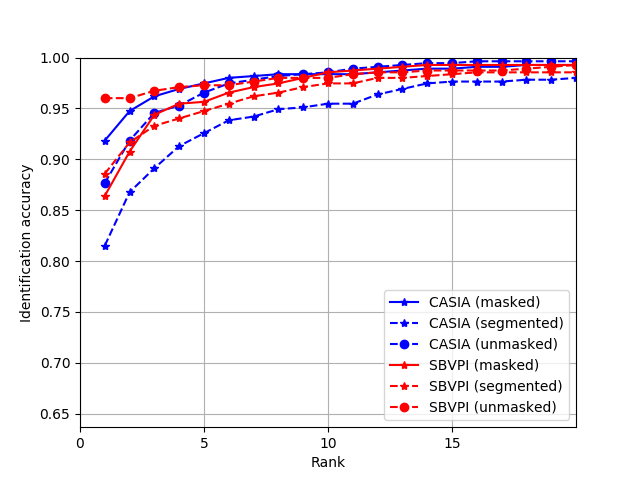}}
\end{minipage}
\hfill
\begin{minipage}{0.325\textwidth}
\centerline{\includegraphics[width=0.995\textwidth,trim={5mm 3mm 15mm 5mm},clip]{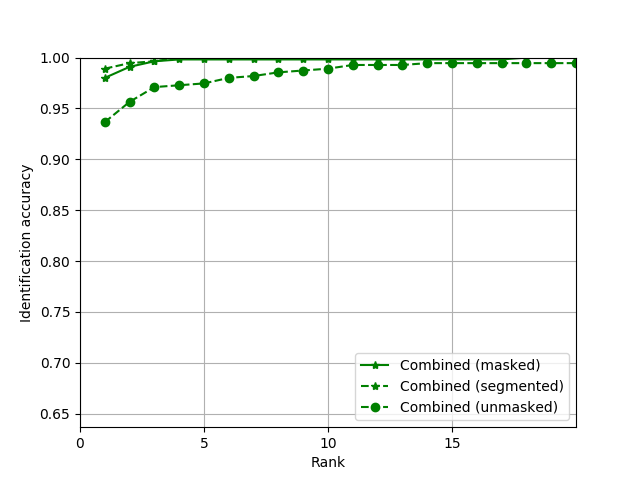}}
\end{minipage}
\hfill
\begin{minipage}{0.325\textwidth}
\centerline{\includegraphics[width=0.995\textwidth,trim={5mm 3mm 15mm 5mm},clip]{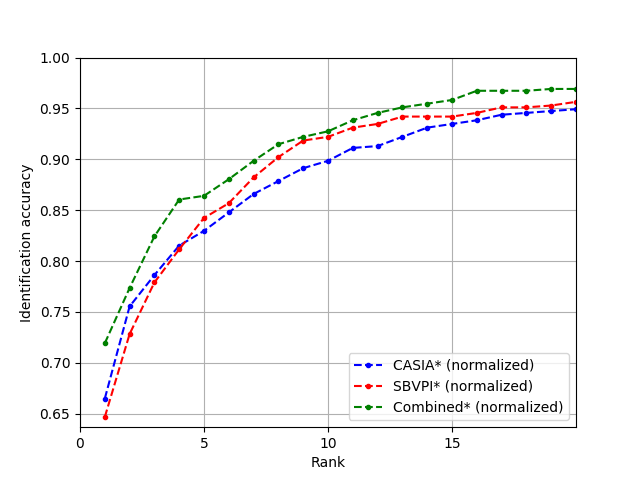}}
\end{minipage}
\caption{CMC curves of models trained on different datasets (CASIA, SBVPI, Combined). Results are shown for different scenarios, where \textit{i)} manually annotated masks are used in the experiments (Masked), \textit{ii)} automatically generated masks are used (Segmented), \textit{iii)} no segmentation masks are used (Unmasked), and \textit{iv)} the iris images are first processed using traditional segmentation techniques and then unwrapped before learning features using a deep iris recognition model (Normalized).}
\label{fig: cmcRec}\vspace{-3mm}
\end{figure*}

To assess the impact of the segmentation procedure on the recognition performance, we conduct four series of experiments using three different training datasets, i.e. CASIA, SBVPI and Combined. In the first series (\textit{Masked} hereafter) we pre-multiply our images with the manually annotated segmentation masks and feed them directly to the recognition model. The idea of this test is to examine the recognition performance with an ideal segmentation mask and compare this to the results achieved with automatically generated segmentation masks in the second series of experiments (marked \textit{Segmented} in the tables). In the third series of experiments, we evaluate the recognition performance without segmentation (referred to as \textit{Unmasked} hereafter). The ``unmasked’’ experiment is used as a baseline in our analysis to evaluate if the segmentation in fact contributes towards better recognition or if using the entire (non-segmented) image (including the periocular part) results in equally good performance. The last series of experiments uses normalized (unwrapped) iris samples. Different from the first three experiments, the recognition model is trained on normalized (square) iris textures instead of masked circular irises. For the normalization procedure we use the iterative Fourier push and pull algorithm from \cite{rathgeb}, which performed best among different models tested in \cite{lozej2018end}. The results of the experiments are presented in Tables \ref{tab: resultsRec}, and \ref{tab: resultsNorm} and plotted in the form of CMC curves in Fig. \ref{fig: cmcRec}.  

\textit{Model trained on CASIA:} The model performs best using the manually annotated segmentation masks and achieves a Rank-1 accuracy of $91.85\%$ in this experiment. This is followed by the results of the unmasked test and finally by the automatic segmentation, which comes in last. We assume that the reason for this is the bad performance of the CASIA segmentation model on samples that are not in CASIA (i.e. samples from SBVPI) as also seen in the third row of Fig. \ref{fig: seg}. Because the CASIA dataset is lacking color information the segmentation isn't able to detect the iris correctly in most cases. We observe that the segmentation works much better in cases where the value and color of the iris is similar as in the greyscale NIR images.

\textit{Model trained on SBVPI:} The model achieves the best Rank-1 score of $96.01\%$ on the unmasked images. Somewhat surprisingly the manual mask proved to have the worst performance, followed by the automatically segmented images with slightly better results. The cause for this may be the diversity in gaze directions present in SBVPI images and the additional difficulty of the color information, which make it difficult to learn discriminative features that generalize well for all variations in image appearance using only so few samples. We assume that with a larger dataset of similar characteristics better results could be achieved. The unmasked samples offer more information based on which the model can construct more general features of the eye as a whole resulting in better performance. From Fig. \ref{fig: seg} (row three) we see that on test samples from the CASIA dataset the segmentation also includes parts of the periocular region, which similarly as in the unmasked experiment may boost performance.

\begin{table}[tb]
\caption{Recognition results for the four experiments. Separate results are shown for each of the three training datasets.}
\begin{center}
\begin{tabular}{|l|c|c|c|}
\hline
\textbf{Approach} & \textbf{Training set} &\textbf{Rank-1 accuracy} & \textbf{Rank-5 accuracy} \\
\hline
Masked & CASIA &\textbf{91.85\%} & \textbf{97.46\%} \\
\hline
Segmented & CASIA & 81.52\% & 92.57\% \\
\hline
Unmasked & CASIA & 87.68\% & 96.56\% \\
\hline
\hline
Masked & SBVPI & 86.41\% & 95.65\% \\
\hline
Segmented & SBVPI & 88.59\% & 94.75\% \\
\hline
Unmasked & SBVPI & \textbf{96.01\%} & \textbf{97.28\%} \\
\hline\hline
Masked  & Combined & 98.01\% & 99.82\% \\
\hline
Segmented & Combined & \textbf{98.91\%} & \textbf{100.00\%} \\
\hline
Unmasked & Combined & 93.66\% & 97.46\% \\
\hline
\multicolumn{3}{l}{In \textbf{bold} - best accuracy for each model}
\end{tabular}
\vspace{-8mm}
\label{tab: resultsRec}
\end{center}
\end{table}

\textit{Model trained on the combined dataset:} This model performs best among all trained models and is the only one able to ensure equally good segmentation and recognition performance on NIR as well as VIS images. Here, the best performance of $98.1\%$ in terms of rank-1 accuracy is achieved by the model using the actual segmentation. From the example (Fig. \ref{fig: seg} – last row) we see that the segmentation results appear better than those of the competing models. The estimated segmentation masks fit tighter to the iris regions and the edges are sharper than those of the other models. The better segmentation performance may be caused by the strong shape information of the CASIA dataset and the diverse color and gaze information, which provided the model with more representative data to train on. Another simpler explanation can be that the increase in performance is because of a larger training dataset. From the visual examples in Fig. \ref{fig: seg}  we also observe that the model doesn't remove small details like thin eyelashes on the surface of the eye, which are removed in the manual annotations. This may also contribute to the performance difference between the Masked and Segmented experiment. Other than that the reason may lie in the joint training of segmentation and recognition. For this model the unmasked results performed worst, although still outperforming the model trained only on CASIA.



\textit{Results for normalized iris textures:} The models trained and tested on normalized iris images preform worst among all models, as seen in Table \ref{tab: resultsNorm} and Fig. \ref{fig: cmcRec}. The main reason for this is that many samples from SBVPI were not normalized correctly (e.g. the iris boundary was not correctly identified), because of the large variability in gaze direction. Several samples also have non-circular irises due to the change in perspective and are therefore harder to segment using traditional techniques. The errors in normalization caused inconsistencies in the normalized samples the result of which is unstable recognition. This is also confirmed by the model trained exclusively on SBVPI, which performs worst out of the three. Still it seems as though the majority of correctly normalized images were also correctly recognized, so we assume that results can be improved using a more effective normalization procedure.

\begin{table}[tb]
\caption{Recognition results with normalized (unwrapped) iris images.}
\begin{center}
\begin{tabular}{|l|c|c|c|}
\hline
\textbf{Approach} & \textbf{Training set} & \textbf{Rank-1 accuracy} & \textbf{Rank-5 accuracy} \\
\hline
\multirow{3}{*}{Normalized} & CASIA* & 66.49\% & 82.97\% \\
\cline{2-4}
& SBVPI* & 64.67\% & 84.23\% \\
\cline{2-4}
& Combined* & 71.92\% & 86.05\% \\
\hline
\multicolumn{4}{l}{In \textbf{*} - Model trained on normalized samples}
\end{tabular}
\label{tab: resultsNorm}\vspace{-5mm}
\end{center}
\end{table}


\section{Conclusion}\label{Sec: Conclusion}
Based on our results we conclude that the use of iris segmentation is important and beneficial in a deep learning setting, under the condition that the segmentation model is able to correctly segment the iris texture from the input images. In our experiments the best results were achieved with a model that incorporated an automatic segmentation step, which achieved even better results than those ensured by manually annotated segmentation masks. 

Another benefit of segmentation, specifically for deep learning, became apparent during training. When training our models on larger datasets (e.g. CASIA-Iris-Thousand) the model only converged correctly (i.e. to a high enough degree), when segmentation was utilized. When we trained directly on the input images the model converged to a low accuracy ($10\%$-$20\%$). So in our case the segmentation also reduced the difficulty of the problem by directing recognition model to a discriminative region,  so training was easier.

We found no advantage in using unwrapped images. In fact, we observed worse results when standard segmentation procedures were used and irises were normalized. The reason, for the worse recognition performance we observed, however, was mostly due to errors in the normalization procedure caused by the non-ideal iris images in the SBVPI dataset. Additional research is needed to further investigate the impact of iris normalization.

\section{Acknowledgment}

Supported in parts by the ARRS Research Programme P2-0250 (B) Metrology and Biometric Systems and P2-0214 (A) Computer Vision, and by the EU through the Horizon 2020 research and innovation program under grants 688201 (M2DC) and 690907 (IDENTITY).
Thanks also to the Nvidia corporation for donating the Titan V GPU.

\bibliographystyle{IEEEtran}
\bibliography{IEEEabrv,bibliography}

\begin{thebibliography}{10}
\providecommand{\url}[1]{#1}
\csname url@samestyle\endcsname
\providecommand{\newblock}{\relax}
\providecommand{\bibinfo}[2]{#2}
\providecommand{\BIBentrySTDinterwordspacing}{\spaceskip=0pt\relax}
\providecommand{\BIBentryALTinterwordstretchfactor}{4}
\providecommand{\BIBentryALTinterwordspacing}{\spaceskip=\fontdimen2\font plus
\BIBentryALTinterwordstretchfactor\fontdimen3\font minus
  \fontdimen4\font\relax}
\providecommand{\BIBforeignlanguage}[2]{{%
\expandafter\ifx\csname l@#1\endcsname\relax
\typeout{** WARNING: IEEEtran.bst: No hyphenation pattern has been}%
\typeout{** loaded for the language `#1'. Using the pattern for}%
\typeout{** the default language instead.}%
\else
\language=\csname l@#1\endcsname
\fi
#2}}
\providecommand{\BIBdecl}{\relax}
\BIBdecl

\bibitem{nigam2015ocular}
I.~Nigam, M.~Vatsa, and R.~Singh, ``Ocular biometrics: A survey of modalities
  and fusion approaches,'' \emph{Infor. Fus.}, vol.~26, pp. 1--35, 2015.

\bibitem{rot2018deep}
P.~Rot, {\v{Z}}.~Emer{\v{s}}i{\v{c}}, V.~Struc, and P.~Peer, ``Deep multi-class
  eye segmentation for ocular biometrics,'' in \emph{IWOBI}.\hskip 1em plus
  0.5em minus 0.4em\relax IEEE, 2018, pp. 1--8.

\bibitem{lozej2018end}
J.~Lozej, B.~Meden, V.~Struc, and P.~Peer, ``End-to-end iris segmentation using
  {U-Net},'' in \emph{IWOBI}.\hskip 1em plus 0.5em minus 0.4em\relax IEEE,
  2018, pp. 1--6.

\bibitem{jalilian2017domain}
E.~Jalilian, A.~Uhl, and R.~Kwitt, ``{Domain Adaptation for CNN Based Iris
  Segmentation},'' \emph{BIOSIG}, 2017.

\bibitem{arsalan2017deep}
M.~Arsalan, H.~G. Hong, R.~Naqvi, M.~B. Lee, M.~C. Kim, D.~S. Kim, C.~S. Kim,
  and K.~R. Park, ``Deep learning-based iris segmentation for iris recognition
  in visible light environment,'' \emph{Symmetry}, vol.~9, no.~11, p. 263,
  2017.

\bibitem{jalilian2017iris}
E.~Jalilian and A.~Uhl, ``Iris segmentation using fully convolutional
  encoder--decoder networks,'' in \emph{Deep Learning for Biometrics}.\hskip
  1em plus 0.5em minus 0.4em\relax Springer, 2017, pp. 133--155.

\bibitem{yang2018robust}
Y.~Yang, P.~Shen, and C.~Chen, ``A robust iris segmentation using fully
  convolutional network with dilated convolutions,'' in \emph{2018 IEEE
  International Symposium on Multimedia (ISM)}.\hskip 1em plus 0.5em minus
  0.4em\relax IEEE, 2018, pp. 9--16.

\bibitem{koh2010robust}
J.~Koh, V.~Govindaraju, and V.~Chaudhary, ``A robust iris localization method
  using an active contour model and hough transform,'' in \emph{ICPR}, 2010,
  pp. 2852--2856.

\bibitem{deepIris}
A.~Gangwar and A.~Joshi, ``{DeepIrisNet}: Deep iris representation with
  applications in iris recognition and cross-sensor iris recognition,'' in
  \emph{IEEE ICIP}, Sept 2016, pp. 2301--2305.

\bibitem{lwIris}
T.~Xingqiang, X.~Jiangtao, and L.~Peihu, ``Deep convolutional features for iris
  recognition,'' in \emph{Biometric Recognition: 12th Chinese Conference, CCBR
  2017, Shenzhen, China, October 28-29, 2017, Proceedings}.\hskip 1em plus
  0.5em minus 0.4em\relax Cham: Springer International Publishing, 2017, pp.
  391--400.

\bibitem{offTheShelf}
K.~Nguyen, C.~Fookes, A.~Ross, and S.~Sridharan, ``Iris recognition with
  off-the-shelf {CNN} features: A deep learning perspective,'' \emph{IEEE
  Access}, vol.~6, pp. 18\,848--18\,855, 2018.

\bibitem{iccv17}
Z.~Zhao and A.~Kumar, ``Towards more accurate iris recognition using deeply
  learned spatially corresponding features,'' in \emph{International Conference
  on Computer Vision, ICCV 2017}, 2017, pp. 1--10.

\bibitem{hofbauer2018exploiting}
H.~Hofbauer, E.~Jalilian, and A.~Uhl, ``{Exploiting Superior CNN-based Iris
  Segmentation for Better Recognition Accuracy},'' \emph{Pat. Rec. Let.}, 2018.

\bibitem{kinnison2019learning}
J.~Kinnison, M.~Trokielewicz, C.~Carballo, A.~Czajka, and W.~Scheirer,
  ``Learning-free iris segmentation revisited: A first step toward fast
  volumetric operation over video samples,'' \emph{arXiv:1901.01575}, 2019.

\bibitem{vgg}
K.~Simonyan and A.~Zisserman, ``Very deep convolutional networks for
  large-scale image recognition,'' \emph{CoRR}, vol. abs/1409.1556, 2014.

\bibitem{incept}
C.~Szegedy, W.~Liu, Y.~Jia, P.~Sermanet, S.~E. Reed, D.~Anguelov, D.~Erhan,
  V.~Vanhoucke, and A.~Rabinovich, ``Going deeper with convolutions,''
  \emph{CoRR}, vol. abs/1409.4842, 2014.

\bibitem{deepLabv3plus}
L.~C. Chen, Y.~Zhu, G.~Papandreou, F.~Schroff, and H.~Adam, ``Encoder-decoder
  with atrous separable convolution for semantic image segmentation,''
  \emph{CoRR}, vol. abs/1802.02611, 2018.

\bibitem{moblienet}
M.~Sandler, A.~G. Howard, M.~Zhu, A.~Zhmoginov, and L.~C. Chen, ``Inverted
  residuals and linear bottlenecks: Mobile networks for classification,
  detection and segmentation,'' \emph{CoRR}, vol. abs/1801.04381, 2018.

\bibitem{xception}
F.~Chollet, ``Xception: Deep learning with depthwise separable convolutions,''
  \emph{CoRR}, vol. abs/1610.02357, 2016.

\bibitem{adam}
D.~P. Kingma and J.~L. Ba, ``Adam: a method for stochastic optimization,'' in
  \emph{ICLR}, 2015, pp. 1--13.

\bibitem{casia}
``{CASIA} {I}ris {I}mage {D}atabase,'' http://biometrics.idealtest.org/.

\bibitem{sbvpi}
``Dataset {SBVPI},'' sclera.fri.uni-lj.si, accessed: 06.10.2018.

\bibitem{rathgeb}
C.~Rathgeb, A.~Uhl, .~Wild, and H.~Hofbauer, ``Design decisions for an iris
  recognition sdk,'' in \emph{Handbook of Iris Recognition}.\hskip 1em plus
  0.5em minus 0.4em\relax Springer, 2016, pp. 359--396.

\end{thebibliography}



\end{document}